\pdfoutput=1
\documentclass[11pt]{article}

\usepackage[]{acl}

\usepackage{times}
\usepackage{latexsym}
\usepackage{graphicx}
\usepackage{enumitem}
\usepackage{multirow}
\usepackage[ruled,linesnumbered]{algorithm2e}

\usepackage[T1]{fontenc}
\usepackage[utf8]{inputenc}

\usepackage{microtype}
\usepackage{smile}

\title{Multilingual Knowledge Graph Completion with Self-Supervised \\ Adaptive Graph Alignment}

\author{Zijie Huang$^{1}$\thanks{~Part of work was done during internship at Amazon; $^{\dagger}$Corresponding author.}, Zheng Li$^{2\dagger}$, Haoming Jiang$^{2}$, Tianyu Cao$^{2}$, Hanqing Lu$^{2}$, \AND Bing Yin$^{2}$, Karthik Subbian$^{2}$, Yizhou Sun$^{1}$, Wei Wang$^{1}$\\
$^{1}$University of California, Los Angeles, CA, USA $^{2}$Amazon.com Inc, CA, USA\\
\texttt{\{zijiehuang,yzsun,weiwang\}@cs.ucla.edu}\\ 
\texttt{\{amzzhe,jhaoming,caoty,luhanqin,alexbyin,ksubbian\}@amazon.com}\\
}

\begin{document}
\maketitle
\begin{abstract}
% \ks{The challenges you describe as not very clear to me. I am not sure whether reviewers would understand. for example, first challenge what do you mean my maximally push parallel entities to be close and ignores KG capacity inconsistency? why do you say prior works are not fully exploited. If some of this already done in language alignment work, why dont we motivate the reader with that. We are inspired by the work done in language alignement world x, y, and z and develop this new approach to handle blah blah blah. Overall, the abstract is not very compelling in the current version.} 
Predicting missing facts in a knowledge graph (KG) is crucial as modern KGs are far from complete. Due to labor-intensive human labeling, this phenomenon deteriorates when handling knowledge represented in various languages. In this paper, we explore multilingual KG completion, which leverages limited seed alignment as a bridge, to embrace the collective knowledge from multiple languages. However, language alignment used in prior works is still not fully exploited: (1) alignment pairs are treated equally to maximally push parallel entities to be close, which ignores KG capacity inconsistency; (2) seed alignment is  scarce and new alignment identification is usually in a noisily unsupervised manner. To tackle these issues, we propose a novel self-supervised adaptive graph alignment (SS-AGA) method. Specifically, SS-AGA fuses all KGs as a whole graph by regarding alignment as a new edge type. As such, information propagation and noise influence across KGs can be adaptively controlled via relation-aware attention weights. Meanwhile, SS-AGA features a new pair generator that dynamically captures potential alignment pairs in a self-supervised paradigm. Extensive experiments on both the public multilingual DBPedia KG and newly-created industrial multilingual E-commerce KG empirically demonstrate the effectiveness of SS-AGA\footnote{Code and data are open-source and available at \url{https://github.com/amzn/ss-aga-kgc}}.

\end{abstract}

 \section{Introduction}
Knowledge graphs (KGs) like Freebase~\cite{freebase} and DBPedia~\cite{DBPedia} are essential for various knowledge-driven applications such as question answering~\cite{qagnn} and commonsense reasoning~\cite{differentiable}. A KG contains structured and semantic information among entities and relations, where prior knowledge can be instantiated as factual triples (head entity, relation, tail entity), e.g., (\textit{Apple Inc.}, \textit{Founded by}, \textit{Steven  Jobs}). As new facts are continually emerging, modern KGs are still far from being complete due to the high cost of human annotation, which spurs on the Knowledge Graph Completion (KGC) task to automatically predict missing triples to complete the knowledge graph.

\begin{figure}[t]
    \centering
  \includegraphics[width=1\linewidth]{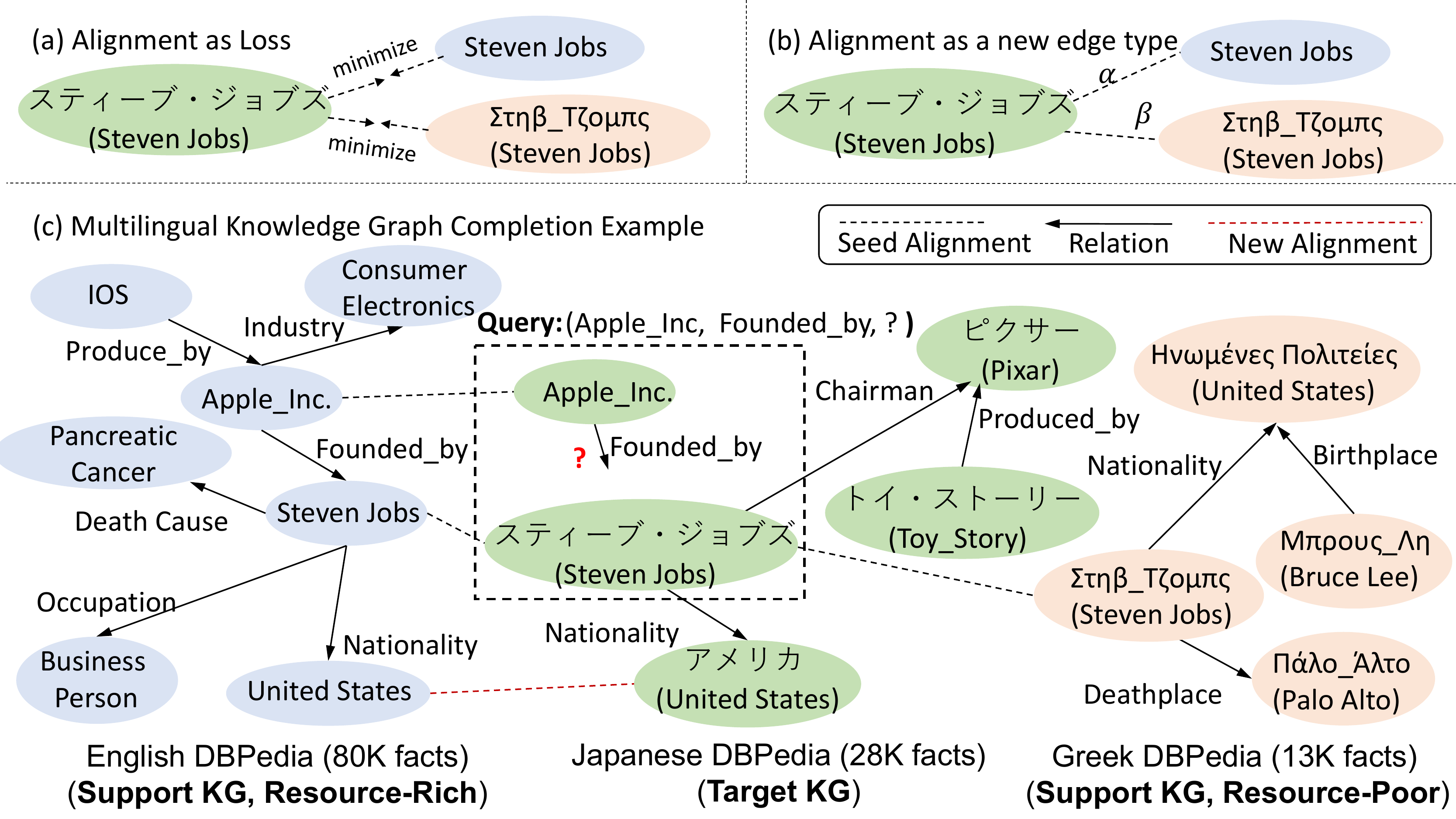}
  \caption{(a) Existing methods treat alignment pairs equally as a loss, which maximally ensures the same entity from different languages to be as similar as possible. (b) Our method differentiates alignment pairs as a new type edge with dynamic attention weights such as $\alpha$ and $\beta$, which control the influence and information propagation from other support KGs. (c) An example of MKGC task answering the query in the Japanese KG.
  }
  \label{fig:motivation_example}
  \vspace{-5mm}
\end{figure}

The KG incompletion circumstance is exacerbated in the multilingual setting, as human annotations are rare and difficult to gather, especially for low-resource languages. Unfortunately, most efforts for KGC have been devoted to learning each monolingual KG separately~\cite{ProcrustEs-KGE,xu-etal-2021-p-int,liang-etal-2021-semantic,cao-etal-2021-missing,lovelace-etal-2021-robust}, which usually underperform in low-resource language KGs that suffer from the sparseness~\cite{mtranse,xuelu,multi_align_survey}. In contrast, KGs from multiple languages are not naturally isolated, which usually share some real-world entities and relations. The transferable knowledge can be treated as a bridge to align different KGs, which not only facilitates the knowledge propagation to low-resource KGs but also alleviates costly manual labeling for all languages.

In this paper, we explore multilingual KG completion (MKGC)~\cite{xuelu} with limited seed alignment across languages. To mitigate language gaps, some efforts have been initiated on multilingual KG embedding methods, which leverage a KG embedding module (e.g., TransE~\cite{transE}) to encode each language-specific KG independently and then employ an alignment loss to force pairs of aligned entities to be close maximally~\cite{xuelu,MultiKE,multi_align_survey}. However, such approaches mainly involve two limitations: (1) the KG inconsistency issue among different languages is neglected due to the equal treatment for parallel entities; (2) the scarcity of seed alignment hinders the efficient knowledge transfer across languages.

% , which can behave as the degrees of incompleteness, knowledge capacity, quality and coverage, etc. 
Concretely, prior methods treat all alignment pairs equally by forcing all parallel entities to be maximally close to each other~\cite{cotrain_text,bootsea,mtranse}. This ignores potentially negative effects from the KG inconsistency due to the language diversity. For example, as shown in Figure~\ref{fig:motivation_example}, the support English KG in DBP-5L~\cite{xuelu} has much more enriched knowledge (80K facts) than the Greek one (13K facts). In order to complete the query (\textit{Apple Inc.}, \textit{Founded by}, ?) in the resource-poor Japanese KG (28K facts), we can transfer more knowledge from resource-rich English KG through the alignment link of \textit{Steven Jobs} than that of the low-data Greek. However, if roughly pushing \textit{Steven Jobs} to be equally close to that English KG and Greek KG, the learned embeddings for \textit{Steven Jobs} will be similar even though they have different structures, KG capacity, coverage and quality. As such, it will bring in irrelevant information regarding this query and may cause the model to get the wrong answer. Thus, we encourage the model to automatically distinguish the underlying inconsistency and transfer knowledge from suitable support KGs\footnote{We regard the remaining KGs as the support KGs when conducting the KGC task in the target one.} for better language-specific KGC performance.

% The suitability can behave as the knowledge capacity, coverage and quality of support KGs with respect to the KGC task of the current language. 

% We have highlighted this in Figure~\ref{fig:motivation_example}(a). Concretely, KGs in different languages may vary in terms of degrees of incompleteness, data size, quality and coverage~\cite{xuelu}. Roughly pushing parallel entities together in the embedding space may strengthen the propagation of unexpected noises or knowledge errors, especially from the low-resource KGs. For example, in Figure~\ref{fig:motivation_example}-(c), English, Japanese and Greek KGs share the same entity \textit{Steven Jobs} with different languages, but the local structure around the entity differs a lot. Besides, the English KG obviously contains more enriched semantics and structural meanings about \textit{Steven Jobs} than that in the Greek KG. If the embedding of the Spanish \textit{Steven Jobs} receives equal influence from the two \textit{Steven Jobs} in English and Greek KGs, the model may be confused whether \textit{Steven Jobs} really founded \textit{Apple Inc.} or not. 

One the other hand, seed alignment is critical for cross-lingual transfer~\cite{xuelu,multi_align_survey}, while acquisition of such parallel entities across languages is costly and often noisy. To mitigate such issue, some recent works~\cite{cotrain_text,xuelu} propose to generate new alignment pairs based on the entity embedding similarity during the training process. The generated new pairs can increase the inter-connectivity between KGs to facilitate knowledge transfer. However, simple usage of correlations between entities without any supervision may increase the noise during training, and inhibit the effectiveness of realistic language alignment in KGs ~\cite{multi_align_survey}.

Motivated by these observations, we propose a \textbf{S}elf-\textbf{S}upervised \textbf{A}daptive \textbf{G}raph \textbf{A}lignment (\textbf{SS-AGA}) 
%\ks{can we call this SAGA? saga means long story of heroic achievements}
framework for MKGC. To tackle the knowledge inconsistency issue, SS-AGA regards alignment as a new edge type between parallel entities instead of a loss constrain, which fuses KGs from different languages as a whole graph. Based on such unified modeling, we propose a novel GNN encoder with a relation-aware attention mechanism, which aggregates local neighborhood information with learnable attention weights and differs the influence received from multiple alignment pairs for the same entity as shown in Figure~\ref{fig:motivation_example}(b). To alleviate the scarcity of seed alignment, SS-AGA exploits a new pair generator that iteratively identifies new alignment pairs in a self-supervised manner. This is achieved by masking some seed alignment in the fused KG before GNN encoding and teaching the generation module to recover them. Empirically, SS-AGA outperforms popular baselines in both public and industrial datasets. For the public dataset, we use the multilingual DBPedia KG~\cite{xuelu} and for the industrial dataset, we create a multilingual E-commerce Product KG called E-PKG. 

Our contributions are as follows: (1) We handle the knowledge inconsistency issue for MKGC by treating entity alignment as a new edge type and introducing a relation-aware attention mechanism to control the knowledge propagation; (2) We propose a new alignment pair generation mechanism with self-supervision to alleviate the scarcity of seed alignment; (3) We constructed a new industrial-level multilingual E-commerce KG dataset; (4) Extensive experiments verify the effectiveness of SS-AGA in both public and industrial datasets.

\section{Preliminaries}

\begin{figure*}[t]
    \centering
   \includegraphics[width=1\linewidth]{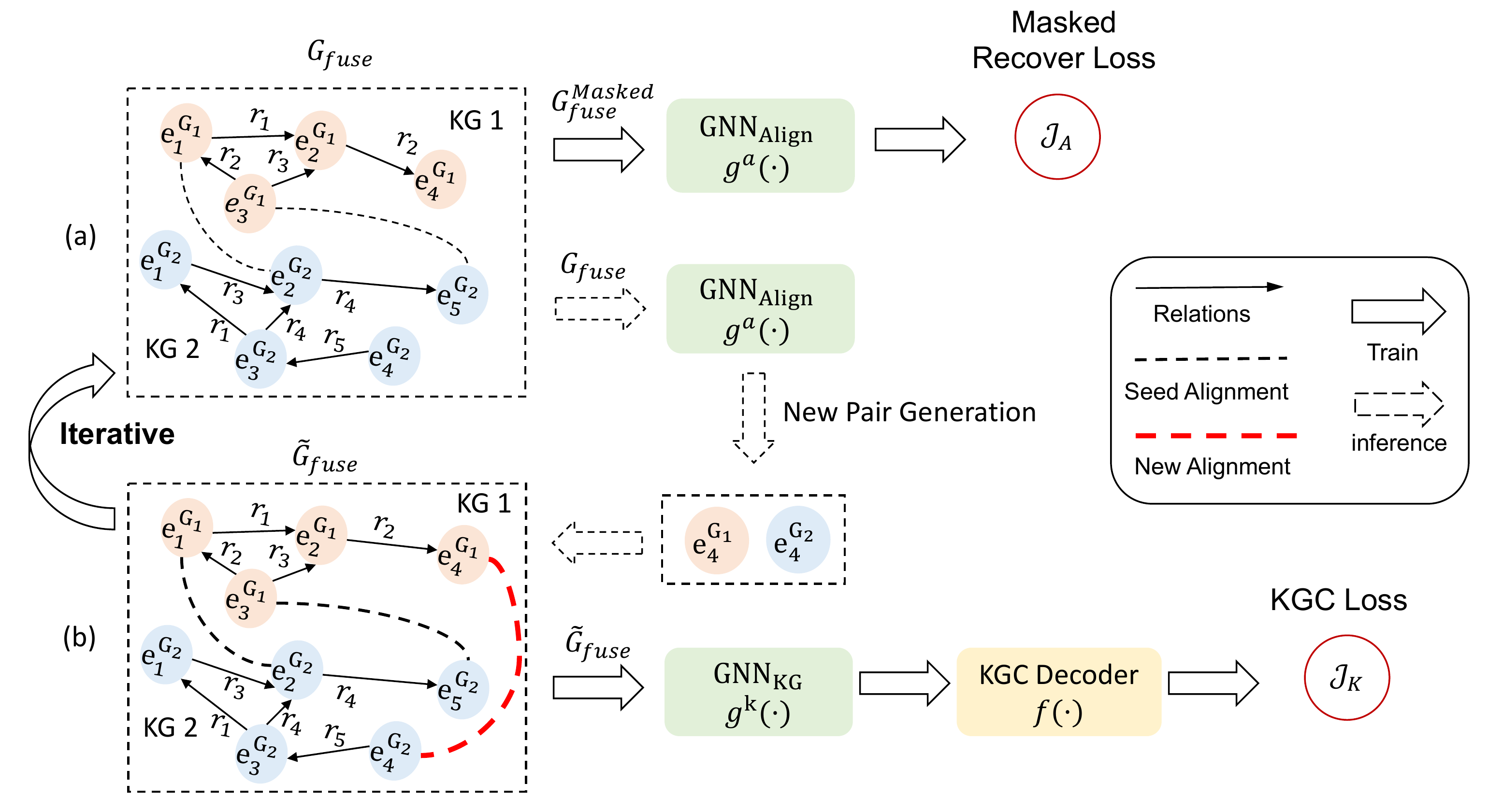}
  \vspace{-2mm}
  \caption{The overall framework of the Self-Supervised Adaptive Graph Alignment (SS-AGA).}
  %Firstly, we fuse different KGs as a whole by introducing cross-KG links based on the alignment information. To propose new alignment pairs during each iteration, (a) we conduct structure learning by masking some alignment links and employing $GNN_{\text{Align}}$ to recover them with a self-supervised learning loss. Then we remove the masked links to generate new pairs from $GNN_{\text{Align}}$. (b) The generated-cross KG links will be added to $G_{\text{fuse}}$. The GNN encoder $GNN_{\text{KG}}$ is employed to compute the contextualized embeddings for each node. Finally, a knowledge graph completion model such as TransE serves as the decoder to compute triple scores.}
  \label{fig:framework}
  \vspace{-5mm}
\end{figure*}

\subsection{Knowledge Graph Completion}

A knowledge graph $G = (\mathcal{E},\mathcal{R}, \mathcal{T})$ consists of a set of entities $\mathcal{E}$, relations $\mathcal{R}$, and relational facts $\mathcal{T}\!=\!\{(e_h,r,e_t)\}$, where $e_h, e_t\!\in\! \mathcal{E}$ are head and tail entities, and $r\!\in\!\mathcal{R}$ is a relation. Entities and relations are represented by their text descriptions. 
The KG completion task seeks to impute the missing head or tail entity of a triple given the relation and the other entity. Without loss of generality, we hereafter discuss the case of predicting missing tails, which we also refer to as a query $q = (e_h,r,?e_t)$.

% We assume that only limited cross-lingual entity alignment pairs are available for knowledge transfer
\noindent\textbf{Multilingual KG completion (MKGC)} utilizes KGs across multiple languages to achieve more accurate KG completion task on each individual KG~\cite{xuelu}. Formally, we are given $M$ different language-specific KGs as $G_1,G_2,\cdots,G_M$, and only limited entity alignment pairs $\Gamma_{G_{i} \leftrightarrow G_{j}} \subseteq \{(e_i,e_j): e_i \in \mathcal{E}_i, e_j \in \mathcal{E}_j\}$ between $G_i$ and $G_j$. We also call $\Gamma_{G_{i} \leftrightarrow G_{j}}$ the \textit{seed} alignment pairs to distinguish it from the new or pseudo alignment.
%A unified relation set is shared across all KGs as similar in~\cite{xuelu}, which are all described in English. 
Each KG $G_{i}$ has their own relation set $\mathcal{R}_{i}$. We denote the union of relation sets from all KGs as a unified relation set $\mathcal{R}\!=\!\mathcal{R}_{1}\!\cup\!\mathcal{R}_{2}\!\cup\!\!\cdots\!\mathcal{R}_{M}$. MKGC is related to but different from the entity alignment (EA) task~\cite{mugnn,multi_align_survey}. In MKGC, seed alignment is not direct supervision while the auxiliary input features, all used in the training stage for cross-lingual transfer to boost the KGC results. 

% The text description of entities $\mathcal{E}_i$ in the $i$-th KG is associated with the $i$-th language, while the relations $\mathcal{R}_i$'s are all described in English. Translating all relations into English is achievable as the number of relations is usually small in most existing knowledge graphs such as DBPedia~\cite{DBPedia} and YAGO~\cite{Yago}.

%We assume the relations are shared across all of them as similar in~\cite{xuelu} as the number of relations is usually much smaller compared to that of entities in many existing knowledge graphs such as DBPedia~\cite{DBPedia}, YAGO~\cite{Yago}, etc. \jhm{what do you mean here}

%, considering that different KGs have their own strengths and limitations on data quality and coverage~\cite{xuelu}. The knowledge transfer is usually achieved by properly utilize the limited entity alignment information among pairs of KGs. 

% Formally, given $M$ different language-specific KGs $G_1,G_2\cdots G_M$, we seek to perform fact prediction on each of those by jointly utilizing the knowledge across all of them, with limited entity alignment information between pairs of KGs. We denote $\Gamma_{G_{i} \leftrightarrow G_{j}}$ as the set of seed entity alignment between KG $G_i$ and $G_j$. We assume the relations are shared across all of them as similar in~\cite{xuelu} as the number of relations is usually much smaller compared to that of entities in many existing knowledge graphs such as DBPedia~\cite{DBPedia}, YAGO~\cite{Yago}, etc.

\subsection{KG Embedding Models}

KG embedding models aim to learn latent low-dimensional representations for entities $\{\be\}_{e \in \cE}$ and relations $\{\br\}_{r \in \cR}$. A naive implementation is an embedding lookup table~\cite{transE,rotatE}. Recently, Graph Neural Networks (GNN) have been explored to aggregate neighborhood information in KGs, where each triple is no longer considered independent of each other~\cite{junheng}. Mathematically, these methods employ a GNN-based encoder $g$ that embeds entities considering the neighborhood information, 
\begin{align*}
    \{\be\}_{e \in \cE} = g(G).
\end{align*}
% They have been widely explored to solve the KGC tasks and 
Then, the plausibility of a relational fact $(e_h,r,e_t)$ can be measured by the triple score:
\begin{align*}
    f(\be_h,\br,\be_t),
\end{align*}
where $f$ can be any scoring function such as TransE~\cite{transE}, RotatE~\cite{rotatE}. We also refer it to as the KGC decoder. 

\section{Method}

%\subsection{Overview}

We introduce SS-AGA for MKGC, consisting of two alternating training components (a) and (b) in Figure~\ref{fig:framework}: (a) A new alignment pair generation module for alleviating the limited seed alignment in $G_{\text{fuse}}$. Specifically, we mask some seed alignment in the fuse KG to obtain $G_{\text{fuse}}^{\text{Masked}}$ and train the generator $g^a(\cdot)$ to recover them. Then, the trained generator will propose new edges based on the learned entity embeddings, which will be incorporated to $G_{\text{fuse}}$ as $\widetilde{G}_{\text{fuse}}$ for MKG embedding model $g^k(\cdot)$ in the next iteration; (b) A novel relation-aware MKG embedding model $g^k(\cdot)$ for addressing the knowledge inconsistency across multilingual KGs. Specifically, we fuse different KGs as a whole graph $G_{\text{fuse}}$ by treating alignment as a new edge type. Then $g^k(\cdot)$ computes the contextualized embeddings for each node with learnable relation-aware attention weights that differ the influence received from multiple alignment pairs. Finally, a KGC decoder $f(\cdot)$ computes the triple scores.

% \zheng{Move to method part. our proposed adaptive GNN-based KG-Embedding model can equip with any kind of existing KGC decoder models like TransE~\cite{transE} and RotatE~\cite{rotatE}. }
% We summarize the {\it SS-AGA} framework in Figure~\ref{fig:framework}

%that uses a novel relation-aware attention-based GNN encoder to learn the contextualized entity embeddings on a fused 
%by adaptively attends to triples from each KG and the alignment links among pairs of KGs. 
%The KG-Embedding Model equips with KGC decoder models to compute triple scores based on the output entity and relation embeddings from the GNN encoder. (2) A new alignment pair generation module that proposes new entity alignment pairs iteratively through a self-supervised learning loss to guide the generation process. The newly-proposed alignment pairs will be used as additional alignment links to augment the input graph structure for the GNN encoder in the next iteration, to update entity and relation embeddings. We summarize the {\it SG-KGE} framework in Figure~\ref{fig:framework}

\subsection{Relation-aware MKG Embedding}
\label{sec:gnn}

As mentioned before, the knowledge transfer is inefficient in existing MKGC methods, as they encode each KG separately and transfer knowledge by forcing aligned entities to share the same embedding. To handle the knowledge inconsistency, we first fuse all KGs as a whole, which relaxes the entity alignment to relational facts. We then design an attention-based relation-aware GNN to learn the contextualized MKG embeddings for entities, which can differ the influence from multiple alignment sources with learnable attention weights. Afterwards, we apply a KGC decoder on the contextualized embedding to get the triple scores for relational facts.

% design a GNN-based KG-Embedding model. 
% design a GNN encoder with an adaptive attention mechanism to embed entities based on a unified fused KG. 

%The goal of the GNN encoder is to conduct efficient knowledge transfer among different KGs and in the meanwhile, handle the knowledge inconsistency issue. We design an attention-based GNN to learn the structural representation for entities across all KGs simultaneously. It therefore embed entities from different KGs into a unified vector space.

%design an adaptive attention mechanism for GNN to learn the structural representation for entities across all KGs simultaneously. 
%It therefore embed entities from different KGs into a unified vector space.

More specifically, we create the fused KG by preserving triples within each KG and converting each cross-KG alignment pair $(e_i,e_j)$ to two relational facts $(e_i,r_{\text{align}}, e_j)$ and $(e_j,r_{\text{align}}, e_i)$ with the alignment edge as a newly introduced relation $r_{\text{align}}$.
%to connect cross-KG entity pairs in the seed alignments. 
%introducing a new type of edge $r_{\text{align}}$ to connect entity pairs that are in the seed alignment as shown in Figure~\ref{fig:framework}. 
% To preserve the symmetry property of alignment, for each pair of aligned entities $e_i, e_j$, we add two triples $(e_i,r_{\text{align}}, e_j)$, $(e_j,r_{\text{align}}, e_i)$ in the fused graph. 
In this way, we enable direct message passing among entities from different KGs, where the attention weight can be learned automatically from data to differ the influence from multiple alignment pairs. We denote the fused knowledge graph as $G_{\text{fuse}}\!=\! (\mathcal{E}_{\text{fuse}},\mathcal{R}_{\text{fuse}}, \mathcal{T}_{\text{fuse}})$, where $\mathcal{E}_{\text{fuse}}\!=\!\bigcup_{i=1}^M \mathcal{E}_i$, $\mathcal{R}_{\text{fuse}}!=\!(\bigcup_{i=1}^M \mathcal{R}_i ) \cup \{r_{\text{align}}\}$ and $\mathcal{T}_{\text{fuse}}\!=\!( \bigcup_{i=1}^M \mathcal{T}_i ) \cup ( \bigcup_{i,j} \{(e_h,r_{\text{align}}, e_t)\!:\! (e_h,e_t) {\rm ~or~} (e_t,e_h) \in \Gamma_{G_{i} \leftrightarrow G_{j}}\}  )$ .

Given the fused KG $G_{\text{fuse}}$, we propose an attention-based relation-aware GNN encoder $g^k(\cdot)$ to learn contextualized embeddings for entities following a multi-layer message passing architecture. 

% At the beginning, the entities and relations are embedded by a lookup table: $e_i \rightarrow \bh^0_i$ and $r \rightarrow \br$.

At the $l$-th layer of GNN, we first compute the relation-aware message delivered by the entity $e_i$ in a relational fact $(e_i,r,e_j)$ as follows:
\vspace{-2mm}
\begin{align*}
    \bh_{i(r)}^l = {\rm Msg}\left(\bh_{i}^l, r\right)  := \bW^l_v {\rm Concat}(\bh_i^l, \br),
        % \label{eq:message}
\end{align*}
where $\bh_i^l$ is the latent representation of $e_i$ at the $l$-th layer, ${\rm Concat}(\cdot,\cdot)$ is the vector concatenation function, and $\bW^l_v$ is a transformation matrix.
% Different from traditional graph attention network \jhm{citation}, $\bh_{i(r)}^l$ captures the relation information by including the relation embedding $\br$.
Then, we propose a relation-aware scaled dot product attention mechanism to characterize the importance of each entity's neighbor $e_i$ to itself $e_j$,
%\footnote{We follow the convention of attention mechanism~\cite{hgt,CG-ODE,LG-ODE} and use ``query entity'' to denote the target entity and ``key entities'' to denote its neighbor entities.}
which is computed as follows: 
% After encoding the relation-aware message, we propose a relation-aware scaled dot product attention mechanism to characterize the importance of each entity's neighbor entities (key entities) to itself (query entity). \footnote{We follow the convention of attention mechanism \jhm{citation} and use ``query entity'' to denote the target entity and ``key entities'' to denote its neighbor entities. }. The neighborhood of a query entity $e_j$ is defined as $\cN(e_j):= \{(e_i,r): (e_i,r,e_j)\in \cT_{\text{fuse}} {\rm ~or~} (e_j,r,e_i)\in \cT_{\text{fuse}} \}$. The attention score between a query entity $e_j$ and a key entity $e_i \in \cN(e_j)$ is computed as follows:
\vspace{-4mm}
\begin{align}
    &{\rm Att}\left(\bh_{i(r)}^l,\bh_j^l\right) = \frac{\exp ( \alpha_{ij}^{r} ) }{\sum_{(e_{i^{\prime}},r)\in \cN(e_j) } \exp \left(\alpha_{i^{\prime}j}^{r}\right)} \notag
    \\
    % \label{eq:softmax} \\
    &\alpha_{ij}^{r} = \left(\bW_k^l \bh_{i(r)}^l \right)^T \cdot \left(\bW_q^l \bh_j^l\right)\cdot \frac{1}{\sqrt{d}}\cdot \beta_r, 
    \label{eq:attention}
\end{align}
where $d$ is the dimension of the entity embeddings, $\bW_k^l, \bW_q^l$ are two transformation matrices, and $\beta_r$ is a learnable relation factor. 
Different from the traditional attention mechanism~\cite{GAT,simgnn}, we introduce $\beta_r$ to characterize the general significance of each relation $r$. It is essential as not all the relationships contribute equally to the query entity. We also remark that the neighborhood is bidirectional, i.e. $\cN(e_j):= \{(e_{i^{\prime}},r): (e_{i^{\prime}},r,e_j)\in \cT_{\text{fuse}} {\rm ~or~} (e_j,r,e_{i^{\prime}})\in \cT_{\text{fuse}} \}$ as the tail entity will also influence the head entity.

We then update the hidden representation of entities by aggregating the message from their neighborhoods based on the attention score:

{\small
\begin{align*}
    \bh_j^{l+1} =  \bh_j^{l} + \sigma\left(\sum_{(e_{i^{\prime}},r)\in \cN(e_j)}{\rm Att}\left(\bh_{{i^{\prime}}(r)}^l,\bh_j^l\right)\cdot \bh_{{i^{\prime}}(r)}^l\right),
    % \label{eq:gnn_aggregate}
\end{align*}}%
where $\sigma(\cdot)$ is a non-linear activation function, and the residual connection is used to improve the stability of GNN~\cite{resnet}.
% One GNN encoder layer only aggregates information from 1-hop neighbors. 

Finally, we stack $L$ layers to aggregate information from multi-hop neighbors and obtain the contextualized embedding for each entity $e_j$ as: $\be_j = \bh_j^L$. Given the contextualized entity embeddings, the KGC decoder computes the triple score for each relational fact: $f(\be_h, \br, \be_t)$. The learning object is to minimize the following hinge loss:

{
\small
\begin{align}
    \mathcal{J}_{K}\hspace{-0.05in}= \hspace{-0.2in}
    % \sum_{m=1}^{M}\hspace{-0.05in}
    % \atop (e_{h^{\prime}}, r, e_{t^{\prime}}) \notin \cT_m
    \sum_{(e_h, r, e_t) \in \cT_m  \atop {(e_{h^{\prime}}, r, e_{t^{\prime}}) \notin \cT_m  \atop m=1,...,M} }\hspace{-0.1in} \left[f\left({\be_h}^{\prime}, \br, \be_t^{\prime}\right)-f\left(\be_h, \br, \be_t\right)+\gamma\right]_{+},
    \label{eq:decoder}
    \end{align}
}%
 where $\gamma>0$ is a positive margin, $f$ is the KGC decoder, $(e_{h^{\prime}}, r, e_{t^{\prime}})$ is a negative sampled triple obtained by replacing either head or tail entity of the true triple $(e_h,r,e_t)$ randomly by other entities in the same language-specific KG.

\begin{remark}
Our method views cross-KG alignment as a relation $r_{\rm align}$ in the fused KG. The knowledge transfer cross KGs is essentially conducted via the learnable attention weight $\alpha_{ij}^{r_{\rm align}}$, where $e_i$ and $e_j$ are connected through the relation $r_{\rm align}$. Thanks to the power of GNN, $\alpha_{ij}^{r_{\rm align}}$ differs the influence from multiple alignment sources, as opposed to some existing models that simply force pairs of entities to be close to each other through a pre-defined alignment loss. In this way, we properly conduct knowledge transfer among KGs with aware of their knowledge inconsistency.
\end{remark}

\noindent\textbf{Scalability issue.} Since we fuse all the $M$ KGs as a whole, and duplicate edges for head entities, the scale of the graph $G_{\text{fuse}}$ would become very large. We therefore employ a $k$-hop graph sampler that samples the $k$-hop neighbors for each node and compute their contextualized embeddings.

\subsection{Self-supervised New Pair Generation}
In multilingual KGs, we are only provided with limited seed alignment pairs to facilitate knowledge transfer, as they are expensive to obtain and even sometimes noisy~\cite{multi_align_survey}. To tackle such challenge, we propose a self-supervised new alignment pair generator. In each iteration, the generator identifies new alignment pairs which will be fed into the GNN encoder $g^{k}(\cdot)$ to produce the contextualized entity embeddings in the next iteration. The training of the generator is conducted in a self-supervised manner, where the generator is required to recover masked alignment pairs. 

\noindent \textbf{New Pair Generation (NPG)} relies on two sets of entity embeddings: the structural embeddings and the textual embeddings. The structural embeddings are obtained by another GNN encoder $g^a$: $\{\be^a\}_{e \in \cE_{\rm fuse}} = g^a(G_{\rm fuse})$, which shares the same architecture with $g^{k}(\cdot)$ in the relation-aware MKG Embedding model (Section~\ref{sec:gnn}). The reason we employ two GNN encoders is that the set of embeddings that generate the best alignment results may differ from those that can best achieve the KG completion task.

The textual embeddings are obtained by entities' text description and mBERT: $\be^{\rm text} = {\rm mBERT}(e)$. mBERT is a multilingual pre-trained language model~\cite{bert} and is particularly attractive to the new alignment pair generation due to the following merits: (1) it captures rich semantic information of the text; (2) the pre-trained BERT embeddings are also aligned across different languages~\cite{bert,multi_align_survey}.

We then model the pairwise similarity score between entity $e_i$ and $e_j$ as the maximum of the cosine similarities of their structural embeddings and textual embeddings:
\begin{align}
    {\rm sim}(e_i,e_j) = \max\left({\rm cos}\left(\be_i^{a},\be_j^{a}\right), {\rm cos}\left(\be_i^{\rm text},\be_j^{\rm text}\right)\right).
    \notag
    % \label{eq:cos}
\end{align}

Then we introduce new alignment pairs if a pair of unaligned entities in two KGs are mutual nearest neighbors according to the cross-domain similarity local scaling (CSLS) measure~\cite{csls} as shown below,

\begin{align}
{\rm CSLS}(e_i, e_j) = 2{\rm sim}(e_i, e_j) - s(e_i) - s(e_j) \notag \\
{\rm subject~to~} s\left(e_i\right)=\frac{1}{K} \sum_{e_{i^{\prime}} \in \mathcal{N}\left(e_i\right)} {\rm sim} \left(e_i, e_{i^{\prime}}\right), \notag
% \label{eq:CSLS}
\end{align}where $K$ is the number of each node's k-nearest neighbors. CSLS is able to capture the sturctural similarity between pairs of entities. The generated pairs are then utilized to update the graph structure of $G_{\text{fuse}}$ to $\widetilde{G}_{\text{fuse}}$ in the next iteration, to alleviate the challenge of limited seed alignment.

\noindent \textbf{Self-Supervised Learning (SSL)} Similar to many existing works~\cite{xuelu,multi_align_survey}, the aforementioned NPG paradigm is unsupervised and may bring in unexpected noises. Inspired by masked language modeling~\cite{bert} which captures contextual dependencies between tokens, we propose a self-supervised learning procedure to guide and denoise the new pair generation. Specifically, we randomly mask out some alignment relational facts, $\cT_{\rm masked} \!\subseteq\! \{(e_h,r,e_t)\!\in\!\cT_{\rm fuse}: r\!=\!r_{\rm align}\}$, and let the generator to recover them. Such masked alignment recovery in KGs can automatically identify the underlying correlations for alignment neighbors and encourage the NPG to generate high-quality alignment pairs that are real existences but hide due to the limited seed alignment.

% This has the same idea as masked language modeling in NLP, which aims to capture contextual dependencies between tokens in the text. 

% self-supervision can teach the generator to distinguish real and non-existent alignment pairs, so as to achieve high-quality alignment. 

Given the fused KG with masked alignment $G_{\text{fuse}}^{\text{Masked}} = \{\cE_{\rm fuse}, \cR_{\rm fuse}, \cT_{\rm fuse}/\cT_{\rm masked}\}$, the GNN encoder $g^a$ embeds the entities as
\begin{align*}
    \{\tilde{\be}\}_{e \in \cE_{\rm fuse}} = g^a(G_{\text{fuse}}^{\text{Masked}}).
\end{align*}

The GNN $g^a$ is then trained via minimizing the following hinge loss $\mathcal{J}_{A}$, 
% \begin{equation}
%     \mathcal{J}_{A} = \sum_{1\leq i < j \leq M} \sum_{\left(e_{i}, e_{j}\right) \in \Gamma_{G_{i} \leftrightarrow G_{j}}^{Masked}}\left\|\tilde{\be}_{i}^{a}-\tilde{\be}_{j}^{a}\right\|_{2}.
%     \label{eq:overall_align_loss}
% \end{equation}
% \sum_{\left(e_{i}, e_{j}\right) \in \Gamma_{G_{i} \leftrightarrow G_{j}}^{Masked},\atop \left(e_{i^{\prime}}, e_{j^{\prime}}\right) \notin \Gamma_{G_{i} \leftrightarrow G_{j}}^{Masked}}
{\small
\begin{align}
\small
\mathcal{J}_{A}^{G_{i} \leftrightarrow G_{j}} &= \hspace{-0.1in}
\sum_{\left(e_{h}, e_{t}\right) \in \Gamma_{ij}^p \atop \left(e_{h'}, e_{t'}\right) \in \Gamma_{ij}^n}\hspace{-0.1in}
\left[\norm{\tilde{\be}_{h}^{a}-\tilde{\be}_{t}^{a}}_2-\norm{\tilde{\be}_{h^{\prime}}^{a}-\tilde{\be}_{t^{\prime}}^{a}}_2+\gamma_{a}\right]_{+} \notag \\
\mathcal{J}_{A} &= \sum_{1\leq i < j \leq M} \mathcal{J}_{A}^{G_{i} \leftrightarrow G_{j}},
    \label{eq:overall_align_loss}
\end{align}
}%
where 
%$dist\left(\tilde{\be}_{i}^{a}-\tilde{\be}_{j}^{a}\right) = \left\|\tilde{\be}_{i}^{a}-\tilde{\be}_{j}^{a}\right\|_{2}$ is the L$2$ distance between entity embeddings, and 
$\Gamma_{ij}^p \hspace{-0.02in}=\hspace{-0.02in} \{(e_h \in \cE_i, e_t \in \cE_j) : (e_h,r_{\rm align},e_t) \in \cT_{\rm masked} \}$ is the masked alignment set, $\Gamma_{ij}^n = \{(e_h \in \cE_i, e_t \in \cE_j) : (e_h,e_t) \notin \Gamma_{G_{i} \leftrightarrow G_{j}} \}$ is the unaligned entity pair set, and
$\gamma_{a} > 0$ is a positive margin. $(e_{h^{\prime}}, e_{t^{\prime}})$ is randomly sampled by replacing one of the entities in the positive entity pairs.

\subsection{Training}
The overall loss function is the combination of the KG completion loss Eq.~\eqref{eq:decoder} and the self-supervised alignment loss Eq.~\eqref{eq:overall_align_loss} as shown below
\begin{equation}
    \mathcal{J} = \mathcal{J}_{K} + \lambda \mathcal{J}_{A},
    \label{eq:overall_loss}
\end{equation}
where $\lambda > 0$ is a positive hyperparameter to balance between the two losses. We summarize the training process in Algorithm~\ref{algo} of the Appendix.

\section{Experiments}
% We present our experimental results to verify the effectiveness of our proposed method. 
% We firstly introduce two real-world datasets that we used, followed by our experimental results and analysis.

%We additionally crawled the text information for entities and relations within each KG. 
% There are two types of text depending on the provided URLs: 1.) text title, which is usually a few words, and 2.) test description, which is usually a detailed paragraph description. Specifically, we crawled text title for all relations. For entities, we crawled text description if they have one, otherwise we crawled text title for them. 
% The alignment information between any pairs of KGs covers around $40\%$ of their total entities. 
% : 

% It is possible for an entity to have multiple alignment pairs across different KG sources. 
\subsection{Dataset}
We conduct experiments over two real-world datasets.  (\romannumeral1) 
\textbf{DBP-5L}~\cite{xuelu} contains five language-specific KGs from DBpedia~\cite{DBPedia}, i.e., English (EN), French (FR), Spanish (ES), Japanese (JA), Greek (EL). As the original dataset only contains structural information, we additionally crawled the text information for these entities and relations based on the given URLs. (\romannumeral2) \textbf{E-PKG} is a new industrial multilingual E-commerce product KG dataset, which describes phone-related product information from an E-commerce platform across six different languages: English (EN), German (DE), French (FR), Japanese (JA),  Spanish (ES),  Italian (IT). The statistics are shown in Table~\ref{table:data_stats}. The $\#$ Aligned Links for a specific KG $G_i$ denotes the number of alignment pairs where one of the aligned entities belong to that KG. It is possible for an entity to have multiple alignment pairs across different KG sources. For both datasets, we randomly split the facts in each KG into three parts: 60$\%$ for training, 30$\%$ for validation, and 10$\%$ for testing. Please refer to {\it  Appendix~\ref{data construct}} for the details of E-PKG construction. 

\begin{table}[htb]
\centering
\resizebox{1\columnwidth}{!}{
\fontsize{8.5}{10.2}\selectfont
\begin{tabular}{@{ }c@{ }|@{ }c@{ }|@{ }c@{ }|@{ }c@{ }|@{ }c@{ }} 
 \toprule
 \hline
    Dataset &\#Entity &\#Relation &\#Triple &\#Aligned Links\\ 
 \hline
 \multicolumn{5}{c}{Multilingual Academic KG ({\it DBP-5L})} \\
 \hline
EN &13,996	&831	&80,167 & 16,916  \\
FR &13,176	&178	&49,015 & 16,877  \\
ES &12,382	&144	&54,066 & 16,347 \\
JA &11,805	&128	&28,774 & 16,263  \\
EL &5,231	&111	&13,839 & 9,042  \\
 \hline
 \multicolumn{5}{c}{Multilingual Industrial KG ({\it E-PKG})} \\
 \hline
EN &16,544  &21  &100,531   & 21,382   \\
DE &17,223  &21  &75,870   & 24,696 \\
FR &17,068  &21  &80,015   & 24,812 \\
JA &2,642  &21  &16,703   & 5,175   \\
ES &9,595  &21  &30,163   & 20,184  \\
IT &15,670  &21  &71,292  & 23,827    \\
 \hline
 \bottomrule
\end{tabular}}
\caption{Statistics of DBP-5L and E-PKG datasets. $\#$Aligned Links denotes the number of alignment pairs where one of the aligned entities belongs to that KG.}
\label{table:data_stats}
\vspace{-5mm}
\end{table}

% \begin{table*}[htbp]
% \centering
% \scalebox{0.85}{
% \begin{tabular}{c|ccccc|cccccc}
% \hline
% \multirow{2}{*}{Dataset} & \multicolumn{5}{c|}{Multilingual Academic KG}  & \multicolumn{6}{c}{Multilingual Industrial KG}    \\   
% \cline{2-12}
% & \multicolumn{5}{c|}{DBP-5L}  & \multicolumn{6}{c}{E-PKG}      
% \\ \hline
% Language            & En    & Fr    & Es    & Jp    & El    & En     & De     & Fr     & Jp    & Es    & It     \\
% \#Entity           & 13996 & 13176 & 12382 & 11805 & 5231  & 16544  & 17223  & 17068  & 2642  & 9595 & 15670  \\
% \#Relation           & 831   & 178   & 144   & 128   & 111   & 21     & 21     & 21     & 21    & 21    & 21     \\
% \#Triples        & 80167 & 49015 & 54066 & 28774 & 13839 & 100531 & 75870 & 80015 & 16703 & 30163 & 71292 \\
% \#Aligned Links & 16917 & 16877 & 16347 & 16263 & 9042  & 21382  & 24696  & 24812  & 5175  & 20184 & 23827  \\ \hline
% \end{tabular}}
% \caption{Statistics of DBP-5L and E-PKG datasets.}
% \label{table:data_stats}
% \end{table*}

\subsection{Evaluation Protocol}
In the testing phase, given each
query $(e_h,r,?e_t)$, we compute the plausibility scores $f(e_h, r,\tilde{e}_t)$ for triples formed
by each possible tail entity $\tilde{e}_t$ in the test candidate set and rank them. We report the mean
reciprocal ranks (MRR), accuracy (Hits@1) and the proportion of correct answers ranked within the top 10 (Hits@10) for testing. 
% All three metrics are preferred to be higher, so as to indicate better KG completion performance.
We also adopt the filtered setting following previous works based on the premise that the candidate space has excluded the triples that have been seen in the training set~\cite{lift1,lift2}. 

\subsection{Baselines}
% We compare our model against both monolingual and multilingual KG completion methods as shown below:

% \begin{itemize}
%     \item \textbf{Monolingual Baselines:} TransE~\cite{transE}, which models relations as translations between head entities and tail entities in a Euclidean space; RotatE~\cite{rotatE}, which models relations as rotations in a complex space; DisMult~\cite{dismult}, a simple bilinear formulation; KG$\-$Bert~\cite{KGbert}, which employs pre-trained language models for knowledge graph completion based on text information of relations and entities. 
%     \item \textbf{Multilingual Baselines:} KEnS~\cite{xuelu}, which jointly captures both the structured knowledge of each KG and the entity alignment that bridges the KGs with ensemble technique;
%     CG$\_$MuAlign~\cite{CG_MuAlign}, which is a GNN-based KG alignment model with specially-designed attention mechanism. We revised the loss function of this model to conduct KG completion task. For both methods, we use mBert~\cite{bert} to generate initial entity  embeddings from their text data, which serves as the initialization value for entity and relation embeddings;
% \end{itemize}

\noindent $\bullet$ \textbf{Monolingual Baselines.}
(\romannumeral1) \textbf{TransE}~\cite{transE} models relations as translations in the Euclidean space; (\romannumeral2) \textbf{RotatE}~\cite{rotatE} models relations as rotations in the complex space; (\romannumeral3) \textbf{DisMult}~\cite{dismult} uses a simple bilinear formulation; (\romannumeral4) \textbf{KG-BERT}~\cite{KGbert} employs pre-trained language models for knowledge graph completion based on text information of relations and entities.

\noindent $\bullet$ \textbf{Multilingual Baselines.}
(\romannumeral1) \textbf{KEnS}~\cite{xuelu} embeds all KGs in a unified space and exploits an ensemble technique to conduct knowledge transfer; (\romannumeral2) \textbf{CG-MuA}~\cite{CG_MuAlign} is a GNN-based KG alignment model with collective aggregation. We revise its loss function to conduct MKGC. (\romannumeral3) \textbf{AlignKGC}~\cite{2021multilingual} jointly trains the KGC loss with entity and relation alignment losses. For fair comparison, we use mBERT~\cite{bert} to obtain initial embeddings of entities and relations from their text for 
all methods. We do not employ any pretrained tasks such as EA to obtain these initial text embeddings as in~\cite{2021multilingual}.

\begin{table}[t]
\centering
\resizebox{0.9\columnwidth}{!}{
\fontsize{8.5}{9.2}\selectfont
\begin{tabular}{@{ }l@{ }|@{ }l@{ }|@{ }c@{ }|@{ }c@{ }|@{ }c@{ }|@{ }c@{ }|@{ }c@{ }} 
 \toprule
\hline
Method                       & Metric & EL            & JA            & ES            & FR            & EN        \\ \hline
 \hline
 \multicolumn{7}{c}{Monolingual Baselines} \\
 \hline
\multirow{3}{*}{TransE}      & H@1     & 13.1          & 21.1          & 13.5          & 17.5          & 7.3             \\
                             & H@10    & 43.7          & 48.5          & 45.0          & 48.8          & 29.3        \\
                             & MRR     & 24.3          & 25.3          & 24.4          & 27.6          & 16.9      \\ \hline
\multirow{3}{*}{RotatE}      & H@1     & 14.5          & 26.4          & 21.2          & 23.2          & 12.3          \\
                             & H@10    & 36.2          & 60.2          & 53.9          & 55.5          & 30.4       \\
                             & MRR     & 26.2          & 39.8          & 33.8          & 35.1          & 20.7      \\ \hline
\multirow{3}{*}{DisMult}     & H@1     & 8.9           & 9.3           & 7.4           & 6.1           & 8.8      \\
                             & H@10    & 11.3          & 27.5          & 22.4          & 23.8          & 30.0       \\
                             & MRR     & 9.8           & 15.8          & 13.2          & 14.5          & 18.3        \\ \hline
\multirow{3}{*}{KG-BERT}     & H@1     & 17.3           &  26.9          &   21.9        & 23.5           &    12.9      \\
                             & H@10    & 40.1          &  59.8         &  54.1         &    55.9      &   31.9     \\
                             & MRR     & 27.3          &   38.7        & 34.0           &   35.4        &   21.0    \\ \hline
\multicolumn{7}{c}{Multilingual Baselines} \\ 
\hline

\multirow{3}{*}{KenS}   & H@1     & 28.1          & 32.1          & 23.6          & 25.5          & 15.1     \\
                             & H@10    & 56.9          & 65.3          & 60.1          & 62.9          & 39.8    \\
                             & MRR     & -             & -             & -             & -             & -         \\ \hline
\multirow{3}{*}{CG-MuA} & H@1     & 21.5          & 27.3          & 22.3          & 24.2          & 13.1        \\
                             & H@10    & 44.8          & 61.1          & 55.4          & 57.1          & 33.5      \\
                             & MRR     & 32.8          & 40.1          & 34.3          & 36.1          & 22.2        \\ \hline 
\multirow{3}{*}{AlignKGC} & H@1     & 27.6          & 31.6          & 24.2          & 24.1          & 15.5        \\
                             & H@10    & 56.3          & 64.3          & 60.9          & 62.3          & 39.2      \\
                             & MRR     & 33.8          & 41.6          & 35.1          & 37.4          & 22.3        \\ \hline
\multirow{3}{*}{\textbf{SS-AGA }}      & H@1     & \textbf{30.8} & \textbf{34.6} & \textbf{25.5} & \textbf{27.1} & \textbf{16.3} \\
                             & H@10    & \textbf{58.6} & \textbf{66.9} & \textbf{61.9} & \textbf{65.5} & \textbf{41.3} \\
                             & MRR     & \textbf{35.3} & \textbf{42.9} & \textbf{36.6} & \textbf{38.4} & \textbf{23.1} \\ \hline
% \multirow{3}{*}{Gain}      & H@1     & 9.6$\%$          & 7.8$\%$           & 8.1$\%$           & 6.3$\%$           & 7.4$\%$          \\
%                              & H@10    & 3.0$\%$          & 2.5$\%$          & 3.0$\%$          & 4.1$\%$           & 3.8$\%$           \\
%                              & MRR     & 7.6$\%$          & 7.0$\%$          & 6.7$\%$           & 6.4$\%$        & 4.1$\%$           \\ \hline
 \bottomrule
\end{tabular}}
\caption{Main results on DBP-5L.}
\label{table:results_dbp5l}
\vspace{-5mm}
\end{table}

\subsection{Main Results}
The main results are shown in Table~\ref{table:results_dbp5l} and Table~\ref{table:results_amazon}. Firstly, by comparing multilingual and monolingual KG models, we can observe that multilingual methods can achieve better performance. This indicates that the intuition behind utilizing multiple KG sources to conduct KG completion is indeed beneficial, compared with inferring each KG independently. Notably, multilingual models tend to bring larger performance gains for those low-resource KGs such as Greek in DBP-5L, which is expected as low-resource KGs are far from complete and efficient external knowledge transfer can bring in potential benefits. Among multilingual models, our proposed method SS-AGA can achieve better performance in most cases across different metrics, languages, and datasets, which verifies the effectiveness of SS-AGA.

\begin{table}[t]
\centering
\resizebox{0.9\columnwidth}{!}{
\fontsize{8.5}{10.5}\selectfont
\begin{tabular}{@{ }l@{ }|@{ }l@{ }|@{ }c@{ }|@{ }c@{ }|@{ }c@{ }|@{ }c@{ }|@{ }c@{ }|@{ }c@{ }}
 \toprule
\hline
Method                       & Metric & EN            & DE            & FR            & JA            & ES & IT    \\ \hline
 \hline
 \multicolumn{8}{c}{Monolingual Baselines} \\
 \hline
\multirow{3}{*}{TransE}      & H@1     & 23.2          & 21.2          & 20.8          & 25.1          & 17.2      & 22.0    \\
                             & H@10    & 67.5          & 65.5          & 66.9          & 72.7          & 58.4     & 63.8  \\
                             & MRR     & 39.4          & 37.4          & 37.5          & 43.6          & 33.0     &  37.8  \\ \hline
\multirow{3}{*}{RotatE}      & H@1     &  24.2         &  22.3         &  22.1         & 26.3          & 18.3     &  22.5  \\
                             & H@10    &  66.8         & 64.3          &  67.1         &  71.9         &  58.9    &  64.0  \\
                             & MRR     &  40.0         & 38.2          &  38.0         &  41.8         & 33.7     &   38.1 \\ \hline
\multirow{3}{*}{DisMult}     & H@1     & 23.8           & 21.4           & 20.7           &  25.9          & 17.9      & 22.8  \\
                             & H@10    & 60.1          & 54.5          & 53.5          & 62.6          & 46.2     & 51.8  \\
                             & MRR     &  37.2          &  35.4         &  35.1         &    38.0      &   30.9   &   34.8 \\ \hline
\multirow{3}{*}{KG-BERT}     & H@1     & 24.3           & 21.8           & 22.3           &  26.9          & 18.7      & 22.9    \\
                             & H@10    & 66.4          & 64.7          & 67.2          &  72.4         & 58.8     & 63.7   \\
                             & MRR     &  39.6          &  38.4         &  38.3         &    44.1      &   33.2   &   37.2 \\ \hline
\multicolumn{8}{c}{Multilingual Baselines} \\ 
\hline
\multirow{3}{*}{KenS}   & H@1     &  26.2        &  24.3         &   25.4        &  33.5         &   \textbf{21.3}     &  \textbf{25.1} \\
                             & H@10    &  69.5         &  65.8         & 68.2         &  73.6         &  59.5    &  \textbf{64.6}  \\
                             & MRR     & -             & -             & -             & -             & -         &-    \\ \hline
\multirow{3}{*}{CG-MuA} & H@1     &   24.8        &  22.9         &  23.0         &  30.4         &  19.2  &  23.9    \\
                             & H@10    &   67.9        &  64.9         &  67.5         &  72.9         &  58.8   &  63.8   \\
                             & MRR     &   40.2        &  38.7         &  39.1           & 45.9          & 33.8   & 37.6  \\ \hline 
\multirow{3}{*}{AlignKGC} & H@1        &   25.6        &  22.1         &  22.8         & 31.2          &19.4      &24.2 \\
                           & H@10      &   68.3        &  65.1         &  67.2         & 72.3          & 59.1     &63.4 \\
                             & MRR     &   40.5        &  38.5         &  38.8         &46.2           &34.2      & 37.3\\ \hline
\multirow{3}{*}{\textbf{SS-AGA}}      & H@1     & \textbf{26.7} & \textbf{24.6} & \textbf{25.9} & \textbf{33.9} & 21.0 & 24.9 \\
                             & H@10    & \textbf{69.8} & \textbf{66.3} & \textbf{68.7} & \textbf{74.1} & \textbf{60.1} & 63.8 \\
                             & MRR     & \textbf{41.5} & \textbf{39.4} & \textbf{40.2} & \textbf{48.3} & \textbf{36.3} & \textbf{38.4}\\ \hline
 \bottomrule
\end{tabular}}
\vspace{-3mm}
\caption{Main results on E-PKG.}
\label{table:results_amazon}
\vspace{-5mm}
\end{table}

\subsection{Ablation Study}
To evaluate the effectiveness of our model design, we conduct ablation study by proposing the following model variants: (\romannumeral1) \textbf{GNN} applies the GNN encoder without relation modeling to each KG independently, and directly forces all alignment pairs to be close to each other as in prior works~\cite{xuelu,CG_MuAlign}; (\romannumeral2) \textbf{R-GNN} is the proposed relation-aware MKG embedding model (Section~\ref{sec:gnn}), which utilizes all seed alignment to construct $G_{\text{fused}}$ and differs the influence from other KGs by the relation-aware attention mechanism; (\romannumeral3) \textbf{R-GNN + NPG} conducts additional new pair generation for R-GNN; (\romannumeral4) \textbf{R-GNN + NPG + SSL} is our proposed full model SS-AGA, which leverages SSL to guide the NPG process. We also investigate the effect of whether to share or not share the encoders $g^a(\cdot),g^k(\cdot)$ that generate the embeddings for the SSL and KGC loss, respectively.

% \textbf{SS-AGA encoder (shared)} shares the parameters of $g^a(\cdot),g^k(\cdot)$ and generates the same embeddings for the SSL and KG completion loss. And \textbf{SS-AGA encoder (no shared)} is our proposed method.

We report the average Hits@1, Hits@10 and MRR over DBP-5L as shown in Table~\ref{table:ablation}. As we can see, applying a GNN encoder to each KG independently would cause the performance drop as all aligned entities are being equally forced to be close to each other. Removing the new pair generation process would also cause a performance degradation due to the sparsity of seed alignment, which shows that iteratively proposing new alignment is indeed helpful. If the generation process is further equipped with supervision, the performance would be enhanced, which verifies the effectiveness of the self-supervised alignment loss. Finally, sharing the parameters of two GNN encoders would harm the performance. Though MKGC and entity alignment are two close-related tasks that can potentially benefit each other, the set of embeddings that produce the best alignment result do not necessarily yield the best performance on the MKGC task.

\begin{table}[htbp]
\centering
\resizebox{1.0\columnwidth}{!}{
\begin{tabular}{llll}
\toprule
\hline
Method                  & Avg H@1 & Avg H@10 & Avg MRR \\ \hline
{GNN} &       24.1       &        56.3
&      33.2        \\
{R-GNN } &    25.7       &        57.9       &      34.4        \\
{R-GNN + NPG}  &    26.2       &        58.3       &      34.9            \\ \hline
\multicolumn{2}{l}{R-GNN + NPG + SSL (SS-AGA)}            \\ 
{ - encoder (shared)} &       25.8       &        57.7
&      34.1  \\
{ - encoder (no shared)} &      \textbf{26.9}       &        \textbf{58.7}         &        \textbf{35.3}          \\ 
 \hline
\bottomrule
\end{tabular}}
\caption{Ablation results on DBP-5L.}
\label{table:ablation}
\vspace{-5mm}
\end{table}

\subsection{Impact of Seed Alignment}
\begin{figure*}[htbp]
    \centering
  \includegraphics[width=1\linewidth]{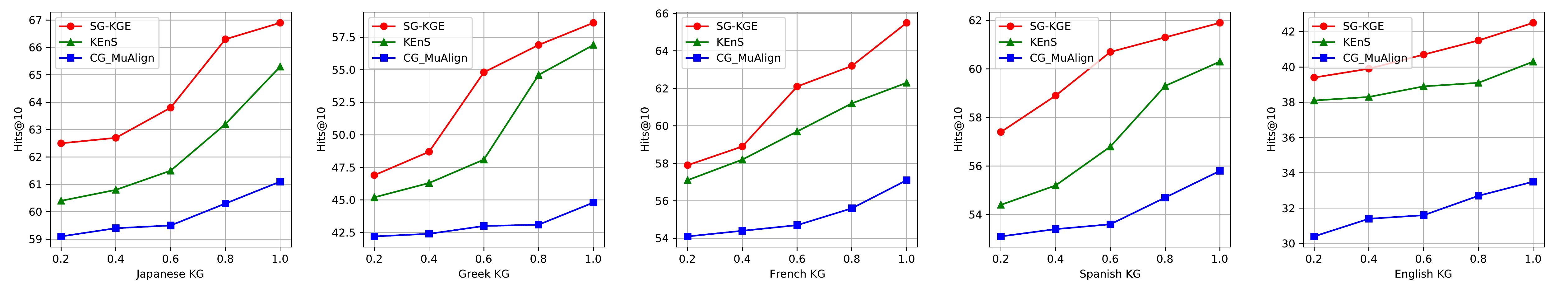}
  \caption{Hits$@$10 with respect to different sampling ratio of seed alignment pairs.}
  \label{fig:seed_alignment}
  \vspace{-6mm}
\end{figure*}

We next study the effect of seed alignment number as depicted in Figure~\ref{fig:seed_alignment}. Firstly, we can observe that SS-AGA consistently outperforms other multilingual models on varying alignment ratios. Secondly, for low-resources KGs such as Japanese and Greek KGs, we can observe a sharp performance drop when decreasing the alignment ratio compared with those popular KGs such as English KG. This indicates that the knowledge transfer among different KGs is especially beneficial for those low-resources KGs, as popular KGs already contain relatively rich knowledge. However, such transfer process is heavily dependent on the seed alignment, which yields the necessity of new alignment generation process.

\subsection{Case Study}

To interpret the knowledge transfer across different KGs, we visualize the normalized average attention weight for each KG w.r.t. the attention score computed in Eq.~(\ref{eq:attention}) from different KG sources. We can see that for those popular KGs, they will receive the highest attention score from themselves such as English and French KGs. Although Japanese KG is low-resource, from the main results table ~\ref{table:results_dbp5l}, we can see that the gap improvement brought by multilingual methods is relatively small compared to another low-resource Greek KG. This indicates that Japanese KG may contain more reliable facts to facilitate missing triple predictions. However, for Greek KG, we can observe that the attention weights from other languages take the majority, which means that the performance boost in Greek KG is largely attributed to the efficient knowledge transfer from other KG sources.
\section{Related Work}
% We firstly discuss related work towards monolingual KG completion and then introduce recent studies on multilingual KG completion.

\subsection{Monolingual KG Embeddings}
Knowledge graph embeddings~\cite{transE,rotatE,ConvE} achieve the state-of-the-art performance for KGC, which learn the latent low-dimensional representations of entities and relations. They measure triple plausibility based on varying score functions such as translation-based TransE~\cite{transE}, TransH~\cite{TransH}; rotation-based RotatE~\cite{rotatE} and language-model-based KG-BERT~\cite{KGbert}.
%TransE views relations as simple translations from a head entity to a tail entity.  
% TransH further extends TransE by projecting entities into relation-specific hyperplanes that enables different roles of an entity in different relations. 
% RotatE~\cite{rotatE} defines relation as rotations in the complex embedding space.
% which is able to capture a lot of useful semantic properties like compositionality of relations. 
% Neural network-based approaches such as ConvE~\cite{ConvE} learn the scoring function along with the model. 
% Some other work~\cite{KGbert} utilizes language models to enhance triple prediction with text input. 
% However, all of the above methods process each triple independently, which to some degree fail to capture the semantically rich neighborhood information. 
Recently, GNN-based methods~\cite{joint_gnn,aaai_gnn,hashtag} have been proposed to capture node neighborhood information for the KGC tasks. GNN is a class of neural networks that operate on graph-structured data by passing local messages~\cite{GCN,GAT,GIN,simgnn,LG-ODE,CG-ODE,DyDiff-VAE}. 
% It has been shown to be powerful for getting contextualized node representations for various graph-related downstream tasks such as node classification~\cite{GCN,GAT}, graph clustering and matching~\cite{simgnn}, etc. 
Specifically, they use GNN as an encoder to generate
contextualized representation of entities by passing local messages~\cite{GCN,GAT,GIN,simgnn,LG-ODE,CG-ODE}. Then, existing score functions are employed to generate triple scores which outperform the aforementioned methods that treat each triple independently only with the scoring function.

\subsection{Multilingual KG Embeddings}
Multilingual KG embeddings are extensions of monolingual KG embeddings that consider knowledge transfer across KGs with the use of limited seed alignment~\cite{multi_align_survey,2021multilingual}. 
%The knowledge transfer relies on the limited seeded alignment entity pairs, whose acquisition is costly and error-prone, especially for multi-lingual KGs.~\cite{multi_align_survey}. 
Earlier work proposes different ways to reconcile KG embeddings for the \textbf{entity alignment (EA)} task: MTransE~\cite{mtranse} learns a transformation matrix between pairs of KGs. MuGNN~\cite{mugnn} reconciles structural differences via rule grounding. CG-MuA utilizes collective aggregation of confident neighborhood~\cite{CG_MuAlign}. Others incorporate attribute information such as entity text~\cite{MultiKE,cotrain_text}. To tackle the sparsity of seed alignment, BootEA~\cite{bootsea} iteratively proposes new aligned pairs via bootstrapping. ~\citet{iterative_entity} 
utilizes parameter sharing  to improve alignment performance. While they focus on the EA     task rather than the MKGC task that we tackle here, such techniques can be leveraged to conduct knowledge transfer among KGs. Recently, \citet{xuelu} propose an ensemble-based approach for the MKGC task. In this paper, we view alignment as a new edge type and employ a relation-aware GNN to get the contextualized representation of entities. As such, the influence of the aligned entities is captured by the learnable attention weight, instead of assuming each alignment pair to have the same impact. We also propose a self-supervised learning task to propose new alignment pairs during each training epoch to overcome the sparsity issue of seed alignment pairs.

\begin{figure}[t]
    \centering
  \includegraphics[width=0.7\columnwidth, height=0.6\columnwidth]{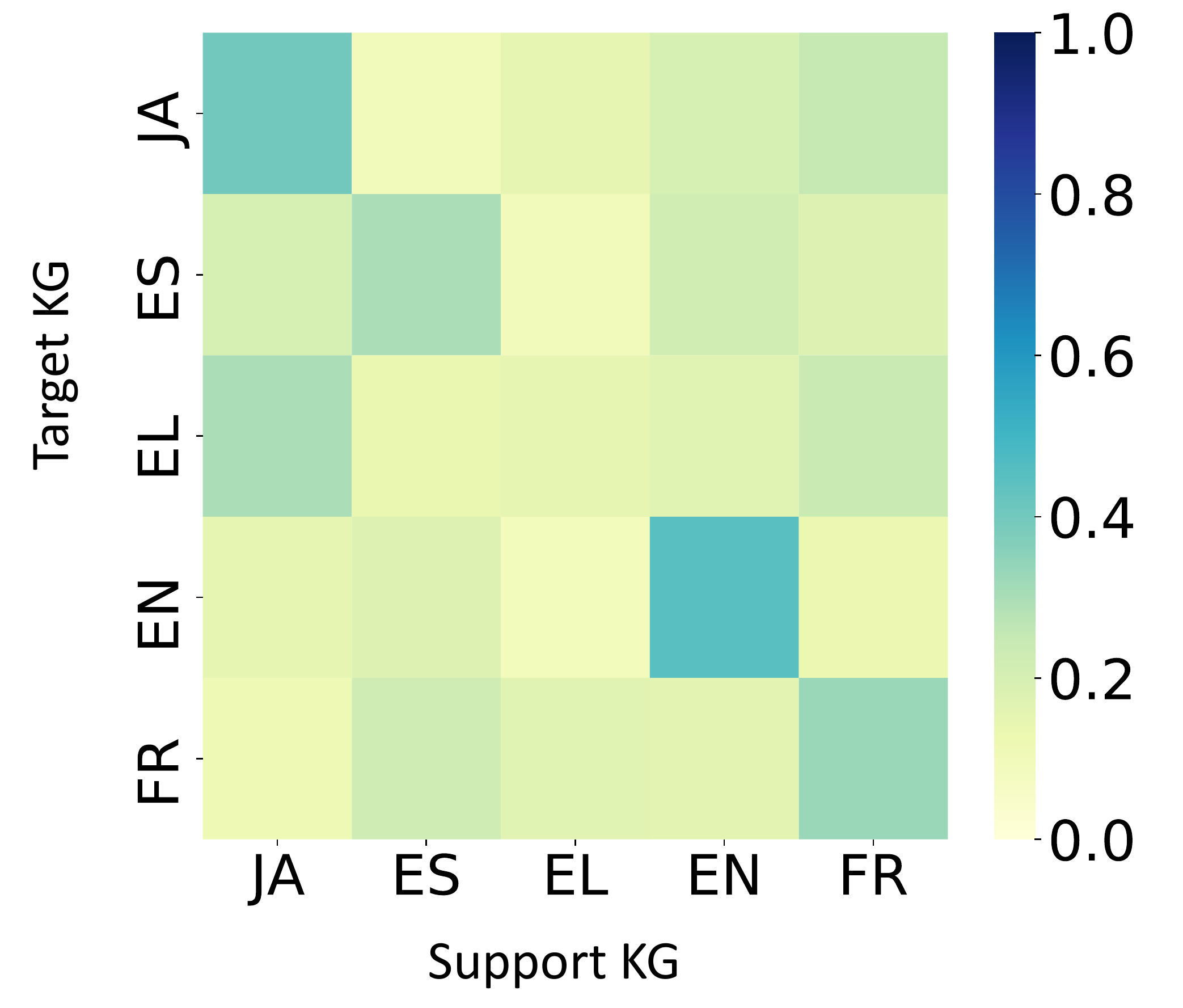}
    \vspace{-2mm}
  \caption{Average attention weight learned in DBP-5L.}
  \label{fig:attention}
  \vspace{-5mm}
\end{figure}
\section{Discussion and Conclusion}
In this paper, we propose SS-AGA for multilingual knowledge graph completion (MKGC). It addresses the knowledge inconsistency issue by fusing all KGs and utilizing a GNN encoder to learn entity embeddings with learnable attention weights that differs the influence from multiple alignment sources. It features a new pair generation conducted in a self-supervised learning manner to tackle the limited seed alignment issue. Extensive results on two real-world datasets including a newly-created E-commerce dataset verified the effectiveness of SS-AGA. Our current approach may fail to fully exploit the benefit of entity and relation texts. In the future, we plan to study more effective ways to combine text data with graph data for better model performance. We are also interested in studying MKGC where there no alignment pairs are given, which is a very practical setting and our current model is not able to deal with.
\section{Ethical Impact}
Our paper proposed SS-AGA, a novel multilingual knowledge graph completion model for predicting missing triples in KGs considering their knowledge transfer. SS-AGA neither introduces any social/ethical bias to the
model nor amplifies any bias in the data. We the created multilingual E-commerce product KG dataset by masking all customers'/sellers' identity and privacy. We only collect information related to products without any personal information leakage. Our model is built upon public libraries in Pytorch. We do not foresee any direct social consequences or ethical issues.

\bibliographystyle{acl_natbib}
\bibliography{custom}

\newpage
\appendix

\section{Data Construction}
\label{data construct}
We introduce the generation process of the multilingual E-commerce KG dataset (E-PKG). E-PKG is a phone-related multilingual product KG across six different languages: English (EN), German (DE), French (FR), Japanese (JA),  Spanish (ES),  Italian (IT). The statistics are shown in Table~\ref{table:data_stats_epkg}. 

\begin{table}[htb]
\centering
\resizebox{1.0\columnwidth}{!}{
\begin{tabular}{l|llllll}
\toprule
\hline
                     & EN      & DE     & FR     & JA     & ES     & IT     \\ \hline
\#Triple\_between    & 90,318  & 65,077 & 69,451 & 14,814 & 23,671 & 60,998 \\
\#Triple\_attributes & 5,013   & 7,345  & 6,017  & 946    & 5,396  & 6,016  \\
\#Triple\_products   & 5,220   & 3,448  & 4,547  & 943    & 1,096  & 4,278  \\
\#Triples            & 100,531 & 75,870 & 80,015 & 16,703 & 30,163 & 71,292 \\
\#Aligned Pairs      & 21,382  & 24,696 & 24,812 & 5,175  & 20,184 & 23,827 \\
\#Entities           & 16,544  & 17,223 & 17,068 & 2,642  & 9,595  & 15,670 \\ 
\#Relations           & 21  & 21 & 21 & 21  & 21 & 21\\ \hline
\bottomrule
\end{tabular}}
\caption{Statistics of E-PKG.}
\label{table:data_stats_epkg}
\end{table}

\begin{figure}[htbp]
    \centering
  \includegraphics[width=1\linewidth]{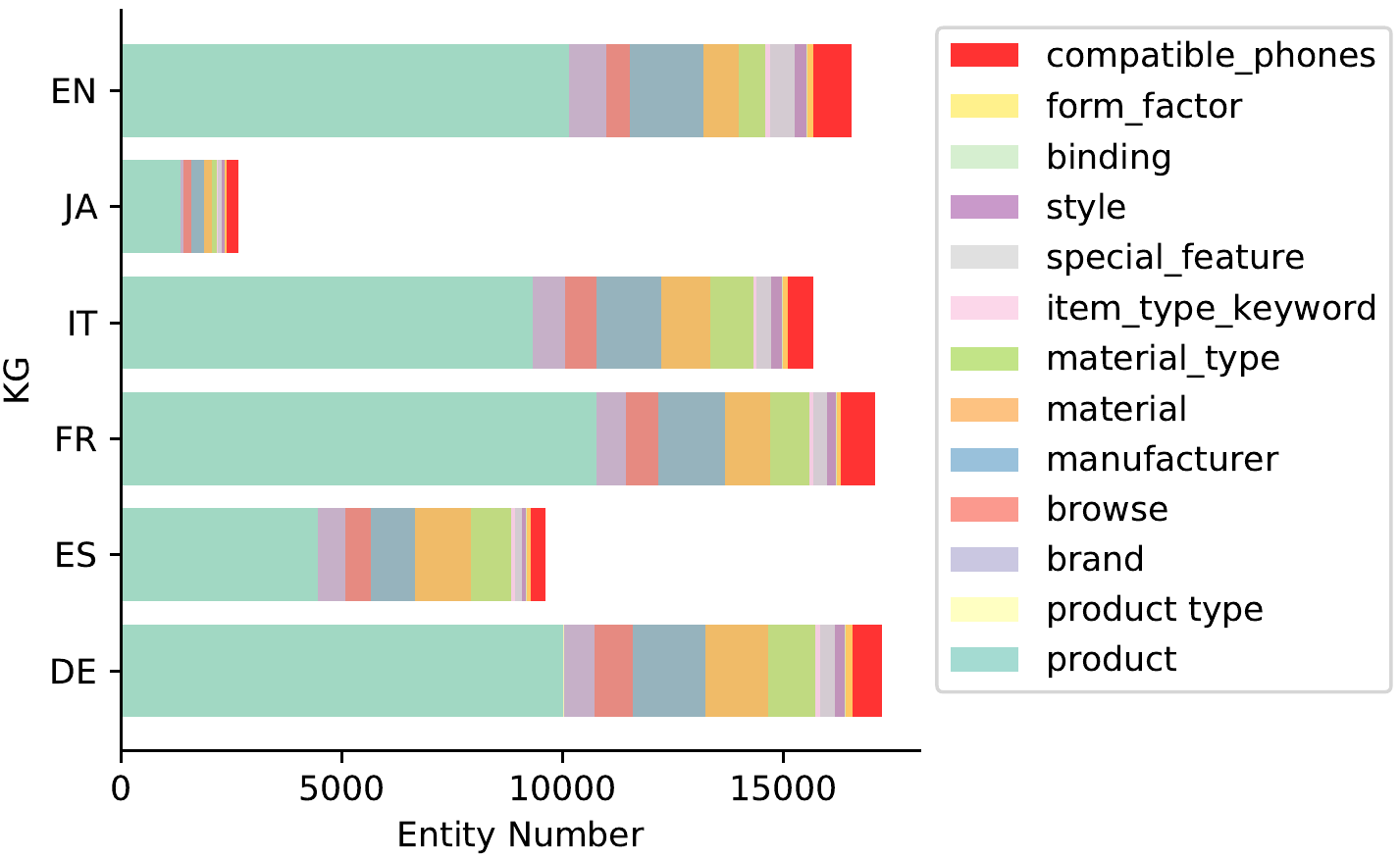}
  \caption{Entity distribution for E-PKG.}
  \label{fig:entity_epkg}
  \vspace{-5mm}
\end{figure}

\begin{figure}[t]
    \centering
  \includegraphics[width=1\linewidth]{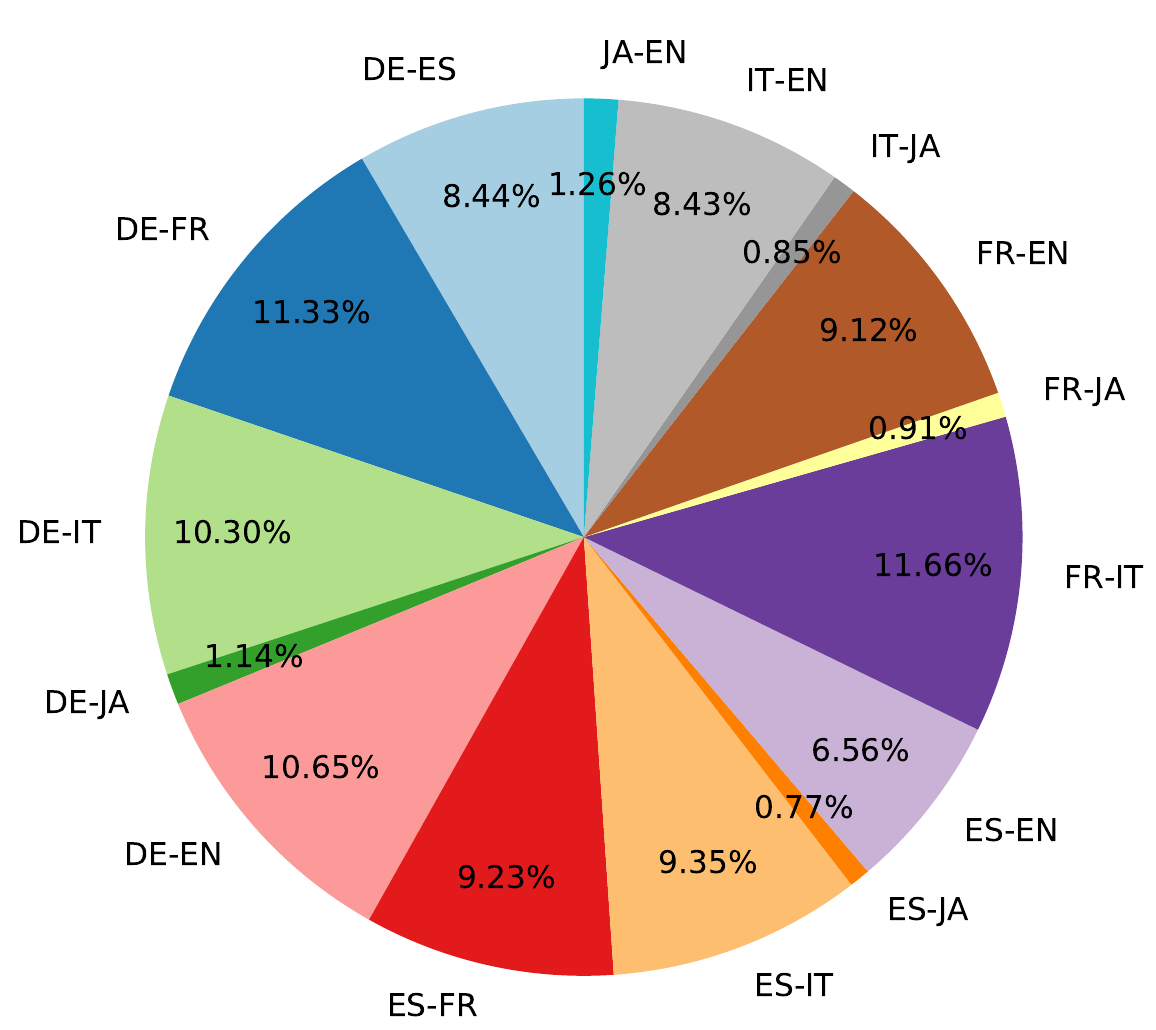}
  \caption{Seed alignment distribution for E-PKG}
  \label{fig:seed_epkg}
  \vspace{-5mm}
\end{figure}

Specifically, each KG consists of two types of entities, which are products such as iPhone 12 and attributes such as style and brand. There are three types of triples grouped by their relation types: 1.) The triples that describe relations between a product and an attribute (Triple$\_$between), such as product-belong-to-brand; 2.) The triples that denote relations between a product and a product, such as product-co-buy-with-product (Triple$\_$products); 3.) The triples that refer to relations between an attribute and an attribute, such as manufacturer-has-brand (Triple$\_$attributes). All relations are described in English and entities are in their own languages. The entity type distributions and seed alignment pairs distributions are illustrated in Figure~\ref{fig:entity_epkg} and Figure~\ref{fig:seed_epkg}, respectively.

\section{Implementation Details}
We use Adam~\cite{adam} as the optimizer to train our model and use TransE~\cite{transE} as the KG decoder whose margin $\gamma$ is set to be 0.3. For the two GNN encoders $g^k(\cdot)$ and $g^a(\cdot)$, we set the latent dimension as 256 with 2 layers, and the dimensions of entity and relation embeddings are also set as 256. We use batch size of 512 and learning rate $lr\!=\!0.005$ during training. The detailed training procedure is illustrated in Algo~\ref{al:training}.
Instead of directly opmizing $\mathcal{J}$ as in Eqn~\ref{eq:overall_loss}, we alternately update $\mathcal{J}_{K}$ and $\mathcal{J}_{A}$ with different learning rate. Specifically, in our implementation, we optimize with $\theta_{new}\leftarrow \theta_{old} - \eta \nabla \mathcal{J}_K$, $\theta_{new}\leftarrow \theta_{old} - (\lambda\eta) \nabla \mathcal{J}_A$ in consecutive steps within one epoch, where $\theta_{new}$ denotes our model parameters and $\nabla$ is the training step.

\begin{algorithm}[htbp]
\caption{SS-AGA training procedure.}
\label{algo}
\label{al:training}
\KwIn{KGs {$G_1,G_2\cdots G_M$};Seed Alignment~{\small$\Gamma_{G_{i} \leftrightarrow G_{j}} (1\leq i<j<M)$}.

\noindent\KwOut{Model parameters $\theta$.}
\While{model not converged}{
//\textit{For the masked alignment pairs}:\\
 Optimize with the masked recover loss in Eqn~\ref{eq:overall_align_loss}:\\
 $\theta_{new}\leftarrow \theta_{old} - (\lambda\eta) \nabla \mathcal{J}_A$\\
//\textit{For new pair generation}:\\
Propose new pairs with all alignment info using CSLS~\cite{csls}\\
//\textit{For KG Completion}:\\
Optimize with the KG completion loss in Eqn~\ref{eq:decoder}:\\
 $\theta_{new}\leftarrow \theta_{old} - \eta \nabla \mathcal{J}_K$\\
    }
}

\end{algorithm}

\end{document}